\documentclass[twoside,11pt]{article}
\usepackage[nohyperref]{jmlr2e}
\usepackage{amsmath}
\usepackage[page]{appendix}
\usepackage{xcolor}
\usepackage[marginparsep=30pt]{geometry}
\usepackage{stmaryrd}
\usepackage{algorithm}
\usepackage{algorithmic}
\usepackage{tikz}
\usepackage{tabu}
\usepackage{listings}
\usepackage{fancyref}
\usepackage{relsize}

\usetikzlibrary{%
    arrows,
    arrows.meta,
    decorations,
    backgrounds,
    positioning,
    fit,
    petri,
    shadows,
    datavisualization.formats.functions,
    calc,
    shapes,
    shapes.multipart,
    matrix
}

\def\version{\texttt{v0.1.0}}
\def\libconform{\texttt{libconform}}

\def\wo{~\\}

\title{\libconform{} \version: a Python library for
       conformal prediction}

\author{\name Jonas Fassbender
        \email jonas@fassbender.dev}

\ShortHeadings{\libconform{} \version{}}{Fa{\ss}bender}
\firstpageno{1}

\begin{document}

\maketitle

\begin{abstract}%
This paper introduces \libconform{} \version{}, a Python
library for the conformal prediction framework, licensed
under the MIT-license.
\libconform{} is not yet stable.
This paper describes the main algorithms implemented and
documents the API of \libconform{}.
Also some details about the implementation and changes in
future versions are described.
\end{abstract}

\begin{keywords}
conformal prediction, Venn prediction, Python,
reliable machine learning
\end{keywords}

\section{Introduction}

This paper introduces the Python library \libconform,
implementing concepts defined in \citet{alrw}, namely the
conformal prediction framework and Venn prediction for
reliable machine learning and predicting with certainty.
These algorithms address a weakness of more traditional
machine learning algorithms which produce only bare
predictions, without their confidence in them/the
probability of the prediction, therefore providing no
measure of likelihood, desirable and even necessary in many
real-world application domains.

The conformal prediction framework is composed of
variations of the conformal prediction algorithm,
first described in
\citet{vovk_et_al_1999, saunders_et_al_1999}.
A conformal predictor provides a measurement of confidence
in its predictions.
A Venn predictor, on the other hand, provides a
multi-probabilistic measurement, making it a
multi-probabilistic predictor.
Below in the text, Venn predictors are included if only
``conformal prediction framework'' is written, except
stated otherwise.

The conformal prediction framework is applied successfully
in many real-world domains, for example face recognition,
medical diagnostic and prognostic and network traffic
classification \citep[see][Part 3]{cprml}.

It is build on traditional machine learning algorithms, the
so called underlying algorithms
\citep[see][]{papadopoulos_et_al_2007}, which makes Python
the first choice for implementation, since its machine
learning libraries are top of the class, still evolving and
improving due to the commitment of a great community of
developers and researchers.

\libconform's aim is to provide an easy to use, but very
extensible API for the conformal prediction framework, so
developers can use their preferred implementations for the
underlying algorithm and can leverage the library, even in
this early stage.
\libconform{} \version{} is \textbf{not} yet stable; there
are still features missing and the API is very likely to
change and improve.
The library is licensed under the MIT-license and its
source code can be downloaded from
\url{https://github.com/jofas/conform}.

This paper combines \libconform{}'s documentation with an
outline of the implemented algorithms.
At the end of each chapter there are notes on the
implementation containing general information about the
library, descriptions of the internal workings and the API
and possible changes in future versions.

Appendix~\ref{appendix:a} provides an overview over
\libconform's API and Appendix~\ref{appendix:b} contains
examples on how to use the library.

\section{Conformal predictors}
\label{sec:cp}

Like stated in the introduction, this chapter will only
outline conformal prediction (CP). For more details see
\citet{alrw}.

\subsection*{Definition}

CP---like the name suggests---determines the label(s) of an
incoming observation based on how well it/they conform(s)
with previous observed examples.

Let $\Lbag z_1,\dots,z_n \Rbag$ be a bag, also called
multiset%
\footnote{It is typical in machine learning to denote this
as a data set, even though examples do not have to
be unique, making the so called set a multiset. A multiset
is not a list, since the ordering of the elements is not
important. If training set, test set, etc.\ is written in
this paper it always denotes a bag, not a set.},
of examples, where each example $z_i \in \textbf{Z}$ is a
tuple $(x_i,y_i); x_i \in \textbf{X}, y_i \in \textbf{Y}$.
\textbf{X} is called the observation space and \textbf{Y}
the label space. For this time \textbf{Y} is considered
finite, making the task of prediction a classification
task, rather than regression, which will be considered in
Chapter~\ref{subsec:rrcm}.

Let $2^\textbf{Y}$ be the set of all subsets of
$\textbf{Y}$, including the empty set.
For example, let $\textbf{Y} := \{0,1\}$;
$2^{\textbf{Y}}$ would be:
\begin{align*}
  \{\{\},\{0\},\{1\},\{0,1\}\}.
\end{align*}

A conformal predictor can be defined as a confidence
predictor $\Gamma$. A confidence predictor is a function
\begin{align*}
\Gamma: \textbf{Z}^* \times \textbf{X} \times (0,1)
\rightarrow 2^\textbf{Y}.
\end{align*}
$\textbf{Z}^*$ denotes a bag of examples with arbitrary
length.

$\Gamma$ takes a bag of examples, a new observation which
should be predicted and $\epsilon \in (0,1)$,
the significance level, as its input and returns the so
called prediction set.
$1 - \epsilon$ is called the confidence level
\citep[see][Chapter 2]{alrw}.

CP produces nested prediction sets.
The prediction sets are called nested, because, for
$\epsilon_1 \geq \epsilon_2$, the prediction set of
$\Gamma^{\epsilon_1}$ is a subset of $\Gamma^{\epsilon_2}$:
\begin{align*}
\Gamma^{\epsilon_1}(\Lbag z_1,\dots,z_n \Rbag, x_{n+l})
\subseteq
\Gamma^{\epsilon_2}(\Lbag z_1,\dots,z_n \Rbag, x_{n+l})
\end{align*}
\citep[see][Chapter 2]{alrw}.

\subsection*{Online vs.\ offline setting}
Conformal prediction can
be used either in the online or the offline---or
batch---learning setting.

In the online setting, after
$\Gamma^{\epsilon}(\Lbag z_1,\dots,z_n \Rbag, x_{n+1})$
has given its prediction, reality would return the
true label $y^{\prime}$ for $x_{n+1}$.
$z_{n+1} := (x_{n+1}, y^{\prime})$ would be appended to
the bag before making the next prediction for $x_{n+2}$.
In the offline setting $x_{n+2}$ would be computed without
$z_{n+1}$ as part of the bag.

The offline learning setting, compared to online learning,
weakens the validity and efficiency---outlined below in
this chapter---of the predictor in favor of computational
efficiency. Also, in many cases the pure online setting is
not even possible or desired, since a predictor giving a
prediction which is then validated from reality directly
afterwards makes the predictor redundant
\citep[see][Chapter 4]{alrw}.

This topic is again discussed below in
Chapter~\ref{sec:icp}.

\subsection*{Transductive vs.\ inductive predictors}

CP was designed to be transductive rather than inductive
\citep[see][Chapter 1]{alrw}.

An inductive predictor $D$ uses a training set
$\Lbag z_1,\dots,z_n \Rbag$ to deduce a decision surface
or prediction rule it uses for predicting a new
observation.

On the other hand a transductive predictor does no such
thing, it rather uses all the previous seen examples from
the training set to predict without deducing a decision
surface beforehand.

While the transductive setting is more elegant than the
inductive setting, it is computationally very expensive and
not feasible for larger bags of examples and for use with
underlying inductive algorithms---discussed in
Chapter~\ref{subsec:ncs}---which have a computationally
complex training phase
\citep[see][Chapter 1]{papadopoulos_et_al_2007, alrw}.

\subsection*{Validity}

A conformal predictor
$\Gamma^\epsilon(\Lbag z_1,\dots,z_n \Rbag, x_{n+1})$
in the online setting is conservatively valid under the
exchangeability assumption.
That means, as long as exchangeability holds, it makes
errors at a frequency of $\epsilon$ or less.
For more on exchangeability and the proof that CP is valid
under exchangeability, refer to
\citet[][Chapters 1--4, 7]{alrw}.

\subsection*{Efficiency}

The efficiency of a conformal predictor can be determined
with many criteria, for example the $N$ criterion.
Let $\Lbag z_{n+1},\dots,z_{n+o} \Rbag$ be a test set.
The $N$ criterion is the average size of the prediction
sets for the test set:
\begin{align*}
  \frac{1}{o}\sum^{n+o}_{i=n+1}|\Gamma^{\epsilon}_i|.
\end{align*}
A small $N$ criteria is preferable
\citep[see][]{vovk_et_al_2016}.

Another criterion would be to determine the frequency of
prediction sets with $|\Gamma^{\epsilon}_i|=1$.
This is an important measure if only predictions with a
single label are desired.

For more efficiency criteria see \citet{vovk_et_al_2016}.

\subsection*{Nonconformity measures}

In order to predict the label of a new observation
$x_{n+1}$,
$\Gamma^{\epsilon}(\Lbag z_1,\dots,z_n \Rbag, x_{n+1})$
sets $z_{n+1}:=(x_{n+1}, y)$, for each $y \in \textbf{Y}$
and checks how $z_{n+1}$ conforms with the examples of the
bag $\Lbag z_1,\dots,z_n \Rbag$.

This is done with a nonconformity measure
$A_{n+1}:\textbf{Z}^n \times \textbf{Z} \rightarrow
\mathbb{R}$. First, $z_{n+1}$ is added to the bag, then
$A_{n+1}$ assigns a numerical score to each example in
$z_i$:
\begin{align}
  \alpha_i = A_{n+1}(\Lbag z_1,\dots,z_{i-1},z_{i+1},
             \dots,z_{n+1} \Rbag, z_i).
\label{eq:a0}
\end{align}
One can see in this equation that $z_i$ is removed from the
bag. It is also possible to compute $\alpha_i$ with $z_i$
in the bag, which means for
$A_{n+1}:\textbf{Z}^{n+1} \times \textbf{Z} \rightarrow
\mathbb{R}$ the score is computed as:
\begin{align}
  \alpha_i = A_{n+1}(\Lbag z_1,\dots,z_{n+1} \Rbag, z_i).
\label{eq:a1}
\end{align}
Which one is preferable is case-dependent
\citep[see][Chapter 4.2.2]{shafer_et_al_2008}.

$\alpha_i$ is called a nonconformity score.

\subsection*{p-values}

The nonconformity scores can now be used to compute the
p-value for $z_{n+1}$, which is the fraction of examples
from the bag which are at least as nonconforming as
$z_{n+1}$:
\begin{align}
  \frac{|\{i=1,\dots,n+1: \alpha_i \geq \alpha_{n+1}\}|}
       {n + 1}.
\label{eq:p0}
\end{align}
If the fraction is close to the upper bound 1 the example
$z_{n+1}$ is very conforming. On the other hand, if it is
close to its lower bound $\frac{1}{n+1}$ it is quite
nonconforming \citep[see][Chapter 2]{alrw}.

Another way to determine the p-value is through smoothing,
in which case the nonconformity scores equal to
$\alpha_{n+1}$ are multiplied by a random value
$\tau_{n+1} \in [0,1]$:
\begin{align}
  \frac{|\{i=1,\dots,n+1: \alpha_i > \alpha_{n+1}\}|
    + \tau_{n+1} |\{i=1,\dots,n+1:\alpha_i=\alpha_{n+1}\}|}
       {n + 1}.
\label{eq:p1}
\end{align}

A conformal predictor using the smoothed p-value is called
a smoothed conformal predictor and is exactly valid under
exchangeability in the online setting, which means it makes
errors at a rate exactly $\epsilon$
\citep[see][Chapter 2]{alrw}.

If the p-value of $z_{n+1}$ is larger than $\epsilon$, $y$
is added to the prediction set.

\begin{algorithm}
  \caption{: Conformal predictor $\Gamma^\epsilon
    (\Lbag z_1,\dots,z_n \Rbag, x_{n+1})$}
  \label{alg:cp}

  \begin{algorithmic}[1]
    \FORALL{$y \in \textbf{Y}$}
      \STATE{set $z_{n+1} := (x_{n+1}, y)$
             and add it to the bag}
      \FORALL{$i=1,\dots,n+1$}
        \STATE{compute $\alpha_i$ with (\ref{eq:a0})
               or (\ref{eq:a1})}
      \ENDFOR
      \STATE{set $p_y$ with (\ref{eq:p0}) or (\ref{eq:p1})}
      \IF{$p_y > \epsilon$}
        \STATE{add $p_y$ to prediction set}
      \ENDIF
    \ENDFOR
    \RETURN{prediction set}
  \end{algorithmic}
\end{algorithm}

\subsection*{Notes on the implementation}

\libconform{} provides the \texttt{CP} class for creating
conformal prediction classifiers. \libconform{}'s
classifier classes provide quite equal APIs, only with
minor variations.
The API of the predictor classes is comparable to major
machine learning libraries like sklearn or keras
\citep[see][]{sklearn_api, keras}.

It is common in machine learning to split the learning task
in two distinct operations, first a
training---or fit---operation on a bag of examples
and then a predict operation on new observations.
\libconform{}'s predictor classes follow this style,
providing a \texttt{train} and a \texttt{predict} method.

While this split in training and predicting is common for
inductive classifiers, which first derive a prediction
rule, or decision surface, from the training set and then
predict unseen examples inductively based on that rule,
it is not really the way CP works. CP was designed to be
transductive, not inductive.

\libconform{}'s aim is to be---one day---ready for
production, where, for some application domains, the time
complexity of predicting a new observation is crucial,
while the time complexity of the training phase
is---in a certain range---not as important.
Therefore \libconform{}'s \texttt{CP} class tries to
minimize the time complexity of its \texttt{predict}
method. Instead of adding $z_{n+1}$ to the bag and then
computing $\alpha_i$ for each example in the bag during
prediction, it computes $\alpha_1,\dots,\alpha_n$ during
training and only computes $\alpha_{n+1}$ in
\texttt{predict} (see Algorithm~\ref{alg:cp}, lines 3--5).
Therefore---not adding $z_{n+1}$ to the bag---it
currently computes the nonconformity scores based on
$A_n$ instead of $A_{n+1}$.

Arguably \texttt{CP} does not implement the conformal
prediction algorithm in its original form (transductive and
online). It provides rather a offline conformal prediction
implementation, or a special case of inductive conformal
prediction, where the calibration set is equal to the whole
bag of examples previously witnessed, instead of a subset
(see Chapter~\ref{sec:icp}).
It is possible that \texttt{CP} will change to being the
implementation of the original conformal prediction
algorithm in a future version, or simply providing an extra
method for the computationally more demanding original
online learning setting \citep[see][Chapter 2]{alrw}.

\texttt{CP} takes an instance of a nonconformity measure
$A$ and a sequence of $\epsilon_1,\dots,\epsilon_{g}$
as its arguments during initialization, therefore being the
implementation of
$\Gamma^{\epsilon_1},\dots,\Gamma^{\epsilon_g}$.

It also provides two extra utility methods for validation,
\texttt{score} and \texttt{score\_online}, which generate
metrics for the conformal predictors
$\Gamma^{\epsilon_1},\dots,\Gamma^{\epsilon_g}$.
The most important of those metrics are the error rates
$Err_1,\dots,Err_{g}$.
It the error rate $Err_i \leq \epsilon_i$ over the bag of
examples provided to \texttt{score}/\texttt{score\_online}
than $\Gamma^{\epsilon_i}$ was valid on the bag.

\texttt{score\_online} adds an example, after it was
predicted, to the training set and calls \texttt{train},
using $\Gamma^{\epsilon_1},\dots,\Gamma^{\epsilon_g}$ in
the online learning setting.

\texttt{CP} also provides a method for another setting of
conformal prediction, this one not based on a significance
level $\epsilon$: \texttt{predict\_best}.
\texttt{predict\_best} always returns a single label, the
one with the highest p-value and optionally also its
significance level. The significance level is the second
highest p-value, since a label is added to the prediction
set---in the original setting---if its p-value is greater
than $\epsilon$ \citep[see][]{papadopoulos_et_al_2007}.

\subsection{Nonconformity measures based on underlying
            algorithms}
\label{subsec:ncs}

Previously nonconformity measures were only described as
any function
$A: \textbf{Z}^* \times \textbf{Z} \rightarrow \mathbb{R}$,
$\textbf{Z}^*$ being any possible bag of examples from
$\textbf{Z}$. This chapter will make a more concrete
description on what nonconformity measures are and how they
use underlying traditional machine learning algorithms.

Let $D:\textbf{Z}^* \times \textbf{X} \rightarrow
\hat{\textbf{Y}}$ be a traditional machine learning
algorithm. $\hat{\textbf{Y}}$ must not be equal to
$\textbf{Y}$. Furthermore there exists a discrepancy
measure $\Delta: \textbf{Y} \times \hat{\textbf{Y}}
\rightarrow \mathbb{R}$.
For a concrete bag $\Lbag z_1,\dots,z_n \Rbag$,
$D_{\Lbag z_1,\dots,z_n \Rbag}$ would be the instance of
$D$ trained on the bag, generating a decision surface based
on it. $D_{\Lbag z_1,\dots,z_n \Rbag}(x)$ would return the
label $\hat{y}$ for $x$. Now we can define the
nonconformity score $\alpha$ for $z := (x, y)$ from the
nonconformity measure $A_n$ as:
\begin{align*}
  \alpha=A_n(\Lbag z_1,\dots,z_n \Rbag, z)=
  \Delta\big(y,D_{\Lbag z_1,\dots,z_n \Rbag}(x)\big),
\end{align*}
or rather with removed example for any $\alpha_i$,
$i = 1,\dots,n$:
\begin{align*}
  \alpha_i = A_n(\Lbag z_1,\dots,z_{i-1},z_{i+1},\dots,z_n
  \Rbag, z_i) = \Delta\big(y_i,D_{\Lbag z_1,\dots,z_{i-1},
  z_{i+1},\dots,z_n \Rbag}(x_i)\big).
\end{align*}

Especially the second equation can be computationally very
complex since it requires to refit $D$ for each
$i = 1,\dots,n$. In general it is not very natural to
use an inductive decision surface $D$ within the
transductive framework of CP.

A popular nonconformity measure is based on the nearest
neighbor method \citep[see][]{alrw,shafer_et_al_2008,cprml,
  smirnov_et_al_2009}.
The general description for the $k$-nearest neighbor method
can be found in \citet{smirnov_et_al_2009}, the other
articles/books describe the nonconformity measure based on
the 1-nearest neighbor method for $z:=(x,y)$ as:
\begin{align*}
  A_n(\Lbag z_1,\dots,z_n \Rbag, z) =
  \frac{\text{min}_{i=1,\dots,n:y_i = y} d(x,x_i)}
       {\text{min}_{i=1,\dots,n:y_i \neq y} d(x,x_i)},
\end{align*}
$d$ being a distance measure, for example the Euclidean
distance. It should be noted that $A_n$ based on the
1-nearest neighbor method for
$A_n(\Lbag z_1,\dots,z_n \Rbag, z_i), i=1,\dots,n$ requires
the removal of $z_i$ from the bag, since otherwise the
smallest distance for $y_i = y_j,j=1,\dots,n$ would always
be 0 resulting in worthless nonconformity scores.

The more general nonconformity measure based on the
$k$-nearest neighbor method can be described as:
\begin{align*}
  A_n(\Lbag z_1,\dots,z_n \Rbag, z)=\frac{d_k^y}{d_k^{-y}},
\end{align*}
$d_k$ being the sum of the $k$ smallest distances to $x$,
$-y$ being all the examples where
$y \neq y_i, i = 1,\dots,n$.

\subsection*{Notes on the implementation}

\libconform{} tries again to be as extensible as possible,
providing a way for developers to define their own
nonconformity measures. For nonconformity measures
\libconform{} provides a module \texttt{ncs} containing
predefined nonconformity measures and a base class for
inheritance called \texttt{NCSBase}.

Predefined are currently the $k$-nearest neighbor method,
one based on a decision tree \citep[see][Chapter 4]{alrw}
and one based on neural networks
\citep[see][Chapter 4]{papadopoulos_et_al_2007,alrw}.
The $k$-nearest neighbor method and the decision tree
are based on the sklearn library
\citep[see][]{sklearn_api}.

Nonconformity measures are classes inheriting from
\texttt{NCSBase} and have to provide an interface with
three methods: \texttt{train}, \texttt{scores} and
\texttt{score}.

\texttt{train}$: \textbf{X}^n \times \textbf{Y}^n$ is for
fitting the underlying algorithm $D$ to a bag of examples.

\texttt{scores}$: \textbf{X}^m \times \textbf{Y}^m \times
bool \rightarrow \mathbb{R}^m$ returning the scores for a
bag of examples. The $bool$ value provided as a parameter
tells the nonconformity measure if the bag is equal to the
bag provided to \texttt{train}, making it possible to
implement (\ref{eq:a0}), rather than (\ref{eq:a1}).
The CP-implementation passes the same bag to \texttt{train}
and \texttt{scores}, while the inductive conformal
prediction implementation (see Chapter~\ref{sec:icp})
passes another bag---the so called calibration set---as a
parameter to \texttt{scores}.

\texttt{score}$: \textbf{X} \times
\textbf{Y}^{|\textbf{Y}|} \rightarrow
\mathbb{R}^{|\textbf{Y}|}$ is for returning the scores of
an example $x$ and each $y \in \textbf{Y}$ combined as
$z := (x, y)$.

\subsection{Conformal predictor for regression: ridge
            regression confidence machine}
\label{subsec:rrcm}

The ridge regression confidence machine algorithm is
described in
\citet[Chapter 2.3]{nouretdinov_et_al_2001, alrw}.
In this chapter, unlike in the previous and following
chapters, $\textbf{Y}$ will be $\mathbb{R}$, making the
prediction a regression problem instead of classification.

Algorithm~\ref{alg:cp} is not feasible for regression,
since $\textbf{Y}$ is now infinite and we would need to
test for each $y \in \textbf{Y}$ if it is in the prediction
set or not. Instead the ridge regression confidence machine
(RRCM) algorithm offers a different approach, returning
prediction intervals instead of prediction sets.

Even though RRCM has ridge regression in its name, it can
be used with other underlying algorithms, like nearest
neighbor regression. For more on ridge regression and its
special case linear regression refer to e.g.
\citet[Chapter 3]{elem_stat}.

Let $\Lbag z_1,\dots,z_n \Rbag$ be our bag of examples,
let $z_{n+1} := (x_{n+1}, y)$ be the observation we want to
predict and let $D_{\Lbag z_1,\dots,z_{n+1} \Rbag}$ be an
underlying regression algorithm.
Previously nonconformity scores were treated as constants,
now we treat them as functions, since $y$ is now an unknown
variable: $\alpha_i(y) = |a_i + b_i y|, i=1,\dots,n+1$.
$a_i$ and $b_i$ are provided by the underlying regression
algorithm. Each $b_i$ is always positive, if not $a_i$ and
$b_i$ are multiplied with $-1$.

Now we can compute the set of $y$'s which p-values are
exceeding a significance level $\epsilon$.
Let $S_i = \{y: |a_i + b_i y| \geq |a_{n+1} + b_{n+1} y|\},
i=1,\dots,n$. Each $S_i$ looks like:
\begin{align*}
S_i =
  \begin{cases}
    [u_i,v_i]
      &\quad \text{if } b_{n+1} > b_i
      \\
    (-\infty, u_i] \cup [v_i, \infty)
      &\quad \text{if } b_{n+1} < b_i
      \\
    [u_i, \infty)
      &\quad \text{if } b_{n+1} = b_i > 0
      \text{ and } a_{n+1} < a_i
      \\
    (-\infty, u_i]
      &\quad \text{if } b_{n+1} = b_i > 0
      \text{ and } a_{n+1} > a_i
      \\
    \mathbb{R}
      &\quad \text{if } b_{n+1} = b_i = 0
      \text{ and } |a_{n+1}| \leq |a_i|
      \\
    \emptyset
      &\quad \text{if } b_{n+1} = b_i = 0
      \text{ and } |a_{n+1}| > |a_i|
  \end{cases},
\end{align*}
so each $S_i$ is either an interval, a point (a special
interval), the union of two rays, a ray, the real line or
empty.
$u_i$ and $v_i$ are either the minimum/maximum of
$-\frac{a_i - a_{n+1}}{b_i - b_{n+1}}$ and
$-\frac{a_i + a_{n+1}}{b_i + b_{n+1}}$,
if $b_{n+1} \neq b_{i}$ or
$u_i = v_i = -\frac{a_i + a_n}{2b_i}$, if
$b_{n+1} = b_i > 0$.
The p-value can only change at $u_i$ or $v_i$.
Therefore all $u_i$ and $v_i$ are sorted in ascending
order generating the sequence $s_1,\dots,s_m$ plus two more
$s$-values, $s_0=-\infty$, $s_{m+1}=\infty$.
The p-value is constant on any interval
$(s_i,s_{i+1}),i=0,\dots,m$ from the sorted set
\citep[see][]{nouretdinov_et_al_2001}.

After that $N$ and $M$ are computed from the sequence.
$N_j, j=0,\dots,m$ for the interval $(s_j,s_{j+1})$ is the
count of $S_i: (s_j,s_{j+1}) \subseteq S_i, i=1,\dots,n$.
$M_j, j=1,\dots,m$, on the other hand, does the same count
only for single $s_j$: $S_i: s_j \in S_i, i=1,\dots,n$.

For a given significance level $\epsilon$ the prediction
interval is the union of intervals from $N$ and points from
$M$ for which $\frac{N_j}{n+1} > \epsilon$ or
$\frac{M_j}{n+1} > \epsilon$, respectively
\citep[see][Chapter 2.3]{alrw}.

In \citet{nouretdinov_et_al_2001} it is stated that there
could be holes in the prediction interval, which means the
the RRCM would return more than a single prediction
interval.
According to the authors these holes rarely show in
empirical tests.
The authors therefore remark that the RRCM can just remove
those holes---therefore returning a single interval---by
simply returning the convex hull of the prediction
intervals.

\subsection*{Notes on the implementation}

The ridge regression confidence machine is implemented as
a class \texttt{RRCM}. It provides the same API as
\texttt{CP}. It implements a computationally less complex
prediction method than the one described above. While the
RRCM described above runs at $\mathcal{O}(n^2)$,
\texttt{RRCM} takes only $\mathcal{O}(n \log n)$, because it
does not compute $N$ and $M$ directly but instead
\begin{align*}
N_{j}^\prime =
  \begin{cases}
    N_j - N_{j-1} &\quad \text{if } j=0,\dots,m \\
    0             &\quad \text{if } j=-1
  \end{cases}
\end{align*}
and
\begin{align*}
M_{j}^\prime =
  \begin{cases}
    M_j - M_{j-1} &\quad \text{if } j=1,\dots,m \\
    0             &\quad \text{if } j=0
  \end{cases},
\end{align*}
which takes only $\mathcal{O}(n)$, making sorting the
$u_i$ and $v_i$ values the most complex task
\citep[see][Chapter 2.3]{alrw}.

The \texttt{RRCM} implementation takes a flag during
its initialization for dealing with the holes described
above, so developers can choose if they want possibly more
than one prediction interval or the convex hull.

\texttt{RRCM} is based on underlying regression algorithms
providing it with its $a_i$ and $b_i$. Currently the
library provides one of these regression algorithms, based
on the $k$-nearest neighbor method.
$a_i$ is the difference between $y_i$ and the average of
the labels of its $k$-nearest neighbors. $y_{n+1}$ is set
to 0, therefore the $k$-nearest neighbor method returns the
negated average of the labels of the $k$-nearest neighbors
of $x_{n+1}$ as $a_{n+1}$.
For $b_i, i=1,\dots,n$ it returns 0, for $b_{n+1}$ it
returns 1.

For developing underlying regression scorers there exists
a base class for inheritance called
\texttt{NCSBaseRegressor}.
It provides a comparable API to \texttt{NCSBase}---the
base class for nonconformity measures---described
in the previous chapter.
Like \texttt{NCSBase} the API contains a \texttt{train}
method for training the underlying algorithm.
Instead of \texttt{scores} and \texttt{score} it has
\texttt{coeffs} and \texttt{coeffs\_n}. The first returns
for a bag two vectors of coefficients $a_i$ and $b_i$ for
each element in the bag. \texttt{coeffs\_n} returns
the coefficients for the observation that needs to be
predicted, in this chapter $z_{n+1} := (x_{n+1},y)$.

\section{Inductive conformal predictors}
\label{sec:icp}

Suppose we have an underlying inductive machine learning
algorithm $D$ as our nonconformity measure and a bag
$\Lbag z_1,\dots,z_n \Rbag$ of examples. If we want to use
$D$ as the nonconformity measure we need to fit it to our
bag: $D_{\Lbag z_1,\dots,z_n \Rbag}$. For some $D$ this can
be a quite time consuming task and in general is not a very
aesthetic thing to do in our transductive setting from the
previous chapter, since---if we would want to predict a
new observation $x_{n+1}$---we would need to refit $D$ for
each $y \in \textbf{Y}$, because we compute our
nonconformity scores adding $z_{n+1} := (x_{n+1},y)$ to
the bag and refitting $D$ with it
(see Algorithm~\ref{alg:cp}, lines 2--5).
Even worse, if we would use (\ref{eq:a0}) instead of
(\ref{eq:a1}) we would need to refit $D$ for each bag
$\Lbag z_1,\dots,z_{i-1},z_{i+1},\dots,z_{n+1} \Rbag,
i=1,\dots,n$.

There exists a natural derivation from the transductive
setting of conformal prediction to the inductive setting
called inductive conformal prediction (ICP).
ICP works more natural with nonconformity measures relying
on inductive machine learning algorithms as the underlying
algorithm \citep[see][Chapter 4]{alrw}.

ICP is computationally less complex than CP, to the cost of
the classifier's validity and efficiency
\citep[see][Chapter 4]{alrw}.

Suppose, again, we have our bag of examples
$\Lbag z_1,\dots,z_n \Rbag$. ICP now splits this bag at a
point $m < n$ into two bags, the so called training set
$\Lbag z_1,\dots,z_m \Rbag$ and the calibration set
$\Lbag z_{m+1},\dots,z_n \Rbag$.

With the training set the underlying algorithm is trained
generating $D_{\Lbag z_1,\dots,z_m \Rbag}$. For each
example in the calibration set the nonconformity score
$\alpha_i$ gets computed:
\begin{align}
  \label{eq:a_icp}
  \alpha_i=\Delta(y_i,D_{\Lbag z_1,\dots,z_m \Rbag}(x_i)),
  i=m+1,\dots,n.
\end{align}

Now, for an incoming example $x_{n+l}$ set
$z_{n+l} := (x_{n+l}, y)$ for each $y \in \textbf{Y}$ and
compute the nonconformity score $\alpha_{n+l}$
like (\ref{eq:a_icp}).
The p-value of $z_{n+l}$ is
\begin{align*}
  \frac{|\{i = m+1,\dots,n: \alpha_i \geq \alpha_{n+l}\}|}
       {n-m+1},
\end{align*}
or the smoothed version
\begin{align*}
  \frac{|\{i=m+1,\dots,n: \alpha_i > \alpha_{n+l}\}|
        + \tau_{n+l}
        |\{i=m+1,\dots,n: \alpha_i = \alpha_{n+l}\}|}
       {n-m+1}
\end{align*}
\citep[see][]{papadopoulos_et_al_2007}.

The huge costs of fitting $D$ repetitively are now reduced
to fitting $D$ only once.
More elaborate update cycles---called teaching
schedules---where $m$ is changing after certain events
and how they impact the validity of the classifier can be
found in \citet[Chapters 4.3, 4.4]{alrw}.

\subsection*{Notes on the implementation}

\texttt{ICP} is the class implementing inductive conformal
prediction. It provides the same API as \texttt{CP},
except \texttt{score\_online}. It has an additional method
\texttt{calibrate} for generating the nonconformity scores
for a the calibration set.

It works with the same nonconformity measures (instances
of classes inheriting from \texttt{NCSBase}) as does
\texttt{CP}.

Currently the nonconformity scores from the calibration set
are saved internally as a vector. In future releases this
will change to an optimized data structure for searching,
e.g.\ a red-black tree \citep[see][]{cormen}.

\section{Mondrian or conditional (inductive) conformal
         predictors}
\label{sec:mcp}

The property of validity under the exchangeablity
assumption can be further optimized with Mondrian or
conditional (inductive) conformal prediction (MCP).
In \citet[Chapter 4.5]{alrw} this form of conformal
prediction is called Mondrian, in \citet[Chapter 2]{cprml}
it is called conditional, the only difference being the
underlying taxonomy, which will be discussed below.

An example from \citet[Chapter 4.5]{alrw} makes it clear
why the stronger form of validity provided by MCP can be
important for some real-world application domains.
The authors tested a 1-nearest neighbor based smoothed
conformal predictor with the significance level
$\epsilon=0.05$ on the USPS data set.
The USPS data set contains 9298 images of handwritten
digits.
The observations are a $16 \times 16$ matrix where each
cell is in the interval of $(-1,1)$ and the labels
obviously are 0 to 9 \citep[see][]{lecun_et_al_1989}.

The authors found out, that while overall the validity held
(the error frequency was nearly equal to $\epsilon=0.05$),
the smoothed conformal predictor had an error rate of
$11.7\%$ on examples with the label ``5''.
The smoothed conformal predictor masked its bad performance
on examples with the label ``5'' simply by predicting other
labels with an error rate of less than $\epsilon = 0.05$,
e.g.\ for the label ``0'' the error rate was below $0.01$
\citep[see][Chapter 4.5]{alrw}.

The idea of MCP is to partition the examples into a
discrete and finite set $\textbf{K}$ of categories
$k \in \textbf{K}$ and to achieve conditional validity in
each category.
For the partitioning a measurable function called a
taxonomy is used.
In \citet[Chapter 4.5]{alrw} the taxonomy is called
Mondrian taxonomy and is defined as:
\begin{align*}
  \kappa: \mathbb{N} \times \textbf{Z} \rightarrow
  \textbf{K},
\end{align*}
in \citet[Chapter 2]{cprml} the taxonomy is called a
$n$-taxonomy:
\begin{align*}
  K_n: \textbf{Z}^n \rightarrow \textbf{K}^n.
\end{align*}

The Mondrian taxonomy $\kappa$ takes the index $i$ of an
example $z_i$ from a sequence $z_1,\dots,z_n$ and $z_i$ as
its input and maps it to a category while the $n$-taxonomy
$K_n$ takes a sequence of examples with a size of $n$ and
maps it to a sequence of categories with size $n$.
$K_n$ is more flexible than $\kappa$ since it is possible
to make the decision which category an example from the
sequence should be in based on the other examples from the
sequence.

The $K$-conditional p-value for an example $z_{n+1}$ and
a bag $\Lbag z_1,\dots,z_n \Rbag$ is now defined for
$i=1,\dots,n+1$ as:
\begin{align}
  \label{eq:mp0}
  \frac{|\{i:K_i = K_{n+1} \text{ \& } \alpha_i
        \geq \alpha_{n+1}\}|}
  {|\{i: K_i = K_{n+1}\}|}.
\end{align}
The smoothed version would be:
\begin{align}
  \label{eq:mp1}
  \frac{|\{i:K_i = K_{n+1} \text{ \& } \alpha_i >
        \alpha_{n+1}\}| + \tau_{n+1}
        |\{i:K_i = K_{n+1} \text{ \& } \alpha_i =
        \alpha_{n+1}\}|}
  {|\{i: K_i = K_{n+1}\}|}.
\end{align}

(\ref{eq:mp0}) and (\ref{eq:mp1}) are the same for $\kappa$
if $K$ is substituted with $\kappa$.

A MCP classifier is category-wise valid under the
exchangeability assumption \citep[see][]{alrw, cprml}.

\subsection*{Notes on the implementation}

There is no direct implementation for MCP, \libconform{}
rather leverages the fact that CP and ICP are just a
special form of Mondrian (inductive) conformal prediction,
where $|\textbf{K}| = 1$, which means all examples are in
the same category.
\texttt{CP} and \texttt{ICP} can take an argument during
initialization called \texttt{mondrian\_taxonomy}.
Currently \texttt{mondrian\_taxonomy} is a function---or
a Python callable rather---which takes one example as its
input and returns the category, basically a 1-taxonomy
$K_1$ where the single example can only be looked at
without context.

In practice a single example often is more information than
really needed.
Often just the label of the example is important, making
the MCP based on this $\textbf{K}$ a label conditional
(inductive) conformal predictor
\citep[see][Chapter 2]{cprml}.

\texttt{mondrian\_taxonomoy} will change in a future
version to $K_n$ for more flexibility.
\\

\noindent
\citet[Chapter 4.5]{alrw} defines Mondrian nonconformity
measures
\begin{align*}
  A: \textbf{K}^* \times \textbf{Z}^* \times
  \textbf{K} \times \textbf{Z} \rightarrow \mathbb{R},
\end{align*}
which add the category to each example in order to compute
the nonconformity scores.
Currently \libconform{} does not have an API for Mondrian
nonconformity measures, which could change in
future releases.

\section{Multi-probabilistic prediction: Venn predictors}
\label{sec:venn}

In the previous chapters we measured the likelihood of a
prediction based on p-values. Even though they produce
valid confidence predictions, the use of p-values is
controversial and they have disadvantages compared to
the probability of a prediction, namely that they are
harder to reason about and that they are often confused
with probabilities \citep[see][Chapter 6.3]{alrw}.

The main negative property of probabilistic prediction is
the fact that it is impossible to estimate true
probabilities---under the unconstrained randomness
assumption---from a finite bag of examples, if the objects
of the bag do not precisely repeat themselves
\citep[see][Chapter 5]{alrw}.

To bypass this property and to achieve a notion of
validity, Venn predictors produce a set of probability
distributions $\{P_y|y\in \textbf{Y}\}$,
$|\textbf{Y}| < \infty$ as their predictions, for which
reason they are called multi-probabilistic predictors
\citep[see][Chapter 2.8]{cprml}.

There are two definitions of validity for a Venn predictor,
the stronger form of validity being that Venn predictors
are ``well-calibrated'' \citep[see][Chapter 6]{alrw}, while
the weaker form of validity states that---under the
unconstrained randomness assumption---a Venn predictor's
prediction is guaranteed to contain the conditional
probability in its multi-probabilistic prediction with
regard to the true probability distribution generating the
examples \citep[see][Chapter 2.8]{cprml}.

A Venn predictor is based on a Venn taxonomy $V_n$ equal to
the $n$-taxonomy described in the previous chapter. Now,
suppose we want to predict the probability distribution for
$z_{n+1} := (x_{n+1}, y); y \in \textbf{Y}$.
We first determine the category $K \in \textbf{K}$ of
$z_{n+1}$ with $V_{n+1}$ and then look at the frequency of
$y$ in this particular category to generate the probability
distribution:
\begin{align*}
P_y = \frac{|\{(x_i,y_i) \in K: y_i = y\}|}{|K|}.
\end{align*}
$K$ is not empty since it at least contains $z_{n+1}$.
This is done for each $y \in \textbf{Y}$ generating the
set of probability distributions
$\{P_y|y \in \textbf{Y}\}$ which is returned as the
multi-probability prediction. The label with the highest
probability is the predicted label.

An example for such a Venn predictor is described in
\citet[Chapter 6.3]{alrw}. It is based on a Venn taxonomy
using the 1-nearest neighbor method to map an example to a
category.
In this case the Venn taxonomy returns the label of the
nearest neighbor as the category.

The Venn predictor generates a matrix
$M:|\textbf{Y}| \times |\textbf{Y}|$. Each example
$z_{n+1} := (x_{n+1},y);y \in \textbf{Y}$ is mapped to a
row.
Each column contains the frequency of $y_i \in \textbf{Y}$
of all examples in the same category as $z_{n+1}$.

The quality of a column is its minimum entry. Now select
the best column $M_{best}$ (with the highest quality) an
return the label of the cell with the highest frequency as
the label prediction and the column as the
multi-probability prediction $\{P_y|y\in \textbf{Y}\}$
\citep[see][Chapter 6.3]{alrw}.

Actually, in the example in \citet[Chapter 6.3]{alrw},
instead of returning $\{P_y|y \in \textbf{Y}\}$ the Venn
predictor returns the interval of the convex hull of the
multi-probability prediction:
$[min\ M_{best},\ max\ M_{best}]$.
This is called the probability interval. The complementary
interval $[1-max\ M_{best},\ 1-min\ M_{best}]$ is called
the error probability interval.

\subsection*{Notes on the implementation}

Venn prediction---like described above---is implemented as
\texttt{Venn}. It again implements the same API as
\texttt{CP} does. The only difference is that
\texttt{Venn}'s \texttt{predict} method takes a flag
\texttt{proba}. If \texttt{proba} is off only the label
prediction is returned. On the other hand, if
\texttt{proba} is set, the label prediction and the error
probability interval is returned.

Currently Venn taxonomies have their own module
\texttt{vtx}. A Venn taxonomy is an instance of a class
that inherits from \texttt{VTXBase}, the same design
like \texttt{NCSBase} and \texttt{NCSBase\-Re\-gres\-sor}
(see Chapter~\ref{sec:cp}).
Once the MCP implementation from the previous chapter moves
from $1$-taxonomies $K_1$ to $n$-taxonomies $K_n$, Venn
taxonomies and $n$-taxonomies will be combined---since they
are equal---and \libconform{} will provide a new API for
both together.

\section{Meta-conformal predictors}
\label{sec:meta}

For understanding how meta-conformal predictors achieve
reliability we have to introduce the concept of abstention
and abstaining classifiers.
An abstaining classifier uses a measure of uncertainty and
if the uncertainty of a new observation is too high the
classifier does not give a prediction, it rejects it
\citep[see][]{vanderlooy_et_al_2009}.

CP is easily modified to produce abstaining classifiers.
We could simply predict the label with the highest
p-value and if its significance level (the second highest
p-value, see Chapter~\ref{sec:cp}) is above a certain
threshold, the classifier returns the label, otherwise it
rejects the observation.

Meta-conformal predictors are described in
\citet{smirnov_et_al_2009}. The authors argue, that if
there exists a classifier $B$ we would need to construct
a nonconformity measure based on $B$; not an easy task,
since there exists no approach for doing so in general.

Therefore they introduce the method of using a conformal
predictor with an already established nonconformity measure
$M$ as a meta-classifier in combination with $B$, so we can
add a certainty measure to a prediction otherwise without
any likelihood indicator.
\textit{B:M} is called the combined classifier.

\subsection*{Meta-classifier instances}

The meta-classifier is a binary classifier that predicts
new observations based on the meta data of the base
classifier $B$. The labels of the meta data are either
0---the negative meta class---if $B$'s prediction
for an observation $x_i$ was wrong or 1---the positive
meta class---if it was correct.

Our meta-conformal predictor $M$ should use a certainty
measure in order to decide for a new observation, if the
prediction of the base classifier $B$ is trustworthy enough
to return. Since $M$ is trained on our meta data with the
positive and negative meta class, $M$ generates two
p-values---one for each class---$p_p$, the positive p-value
and $p_n$, the negative p-value.
We can use both p-values to convert \textit{B:M} to a
scoring classifier with a score ratio $\frac{p_p}{p_n}$.
Now we only need to define the reliability threshold $T$.
If the score ratio generated by $M$ for a new observation
is greater than $T$ we say the prediction of $B$ is
trustworthy and is returned, otherwise \textit{B:M}
abstains from making the prediction
\citep[see][]{smirnov_et_al_2009}.

\subsection*{Performance metrics for binary classifiers}

Metrics for the performance of a binary classifier can
be given by a confusion matrix (see Figure~\ref{fig:cm}).
For a test set the confusion matrix counts the predicted
examples and maps them---depending on the true class and
the predicted class---to its entries.

\begin{figure}
\begin{center}
\begin{tikzpicture}
  \def\th{1em}
  \def\tw{4.2cm}
  \def\tww{1.5cm}
  \def\twww{1.65cm}

  \node[ rectangle split
       , rectangle split parts = 3
       , text width = \tw, text height = \th
       , draw, label=positive ] (a) at (0,0)
  {%
    True Positive ($TP$)
    \nodepart{two} False Negative ($FN$)
    \nodepart{three} Rejected Positive ($RP$)
  };

  \node[ right=0 of a.east
       , rectangle split
       , rectangle split parts = 3
       , text width = \tw, text height = \th
       , draw, label=negative ]
  {%
    False Positive ($FP$)
    \nodepart{two} True Negative ($TN$)
    \nodepart{three} Rejected Negative ($RN$)
  };

  \node[above=2em of a.north east]{True class};

  \matrix[left=0 of a.170, left delimiter=\{] (m)
  {%
    \node[text width=\tww,text height=\th]{positive};\\
    \node[text width=\tww,text height=\th]{negative};\\
  };

  \node[left=0 of m.west, text width=2cm] {Predicted class};

  \node[
    left=0 of a.196, text width=\twww, text height=\th
  ] {rejected};
\end{tikzpicture}
\caption{Confusion matrix for binary abstention
         classifiers.}
\label{fig:cm}
\end{center}
\end{figure}

From the confusion matrix we can derive interesting metrics
for the binary classifier, the most important being the
accuracy $A$, the precision rate $P$:
\begin{align*}
A = \frac{TP + TN}{TP + TN + FP + FN};\
P = \frac{TP}{TP + FP},
\end{align*}
the true positive rate $TPr$ and the false positive rate
$FPr$:
\begin{align*}
TPr = \frac{TP}{TP + FN},\ FPr = \frac{FP}{FP + TN}.
\end{align*}

For a binary abstention classifier we can also add the
rejection rate
\begin{align*}
R = \frac{RP + RN}{RP + RU + TP + TN + FP + FN}.
\end{align*}

\subsection*{Constructing the combined classifier
             \textit{B:M}}

In order to construct a combined classifier \textit{B:M}
we first need the meta data.
For that we use the $k$-fold method normally used as a
cross-validation technique
\citep[see][Chapter 7.10]{elem_stat}.

Let $\Lbag z_1,\dots,z_n \Rbag,\ z_i \in \textbf{Z}:
\textbf{X} \times \textbf{Y}$ be our training set.
We split the training set in $k$ roughly equal sized
partitions.
For each partition: take the partition as test set,
combine the others to a training set and fit $B$ to it.
Let $B$ predict on the test set which then generates a
partition of the meta data
$\Lbag z^{\prime}_i,\dots,z^{\prime}_{i+l} \Rbag,\
z^{\prime}_i \in \textbf{Z}^{\prime}: \textbf{X} \times
\{0,1\}$ (see Algorithm~\ref{alg:kfoldmeta}).

After we are through with Algorithm~\ref{alg:kfoldmeta} and
we have generated our meta data
$\Lbag z^{\prime}_1,\dots,z^{\prime}_n \Rbag$, we
potentially could train $M$ to it, but we still have to
define the reliability threshold $T$.

Having a target accuracy $A_t$ \citet{smirnov_et_al_2009}
proposes to use a ROC isometrics approach defined in
\citet{vanderlooy_et_al_2009} to set $T$ based on $A_t$.
In the case where we have a combined classifier
\textit{B:M} $A_t$ equals the precision rate $P_M$ of $M$
\citep[see][]{smirnov_et_al_2009}.

Defining $T$ based on $P_M$ is done in five steps:

\begin{enumerate}

  \item use the same $k$-fold algorithm used to generate
        the meta data---with $M$ instead of $B$---to
        generate a set of scoring ratios $\frac{p_p}{p_n}$.

  \item construct the ROC curve based on the $TPr$ and
        $FPr$ of $M$ which indirectly maps to the scoring
        ratios. For more information about ROC curves
        see e.g. \citet{fawcett_2006}.

  \item abstract the convex hull ROCCH of the ROC curve.

  \item construct the iso-precision line from the equation
        $TPr = \frac{P_M}{1-P_M} \frac{N}{P} FPr$, $N$
        being the number of negative meta instances, $P$
        being the number of positive meta instances.
        This line represents all classifiers with a target
        precision rate of $P_M$. The classifier on the line
        which is abstaining the least is determined in the
        next step.

  \item set $T$ as the scoring ratio $\frac{p_p}{p_n}$ at
        the intersection of the ROCCH and the iso-precision
        line.

\end{enumerate}

After we have determined $T$, $B$ is trained on the whole
training set $\Lbag z_1,\dots,z_n \Rbag$, $M$ on the
whole meta data set
$\Lbag z^{\prime}_1,\dots,z^{\prime}_n \Rbag$ and the
training operation of \textit{B:M} is over.

\begin{algorithm}
  \caption{: k-fold method for meta data generation}
  \label{alg:kfoldmeta}

  \textbf{Input:}

  \quad $B$: a classifier,

  \quad bag: a bag of examples $\Lbag z_1,\dots,z_n \Rbag$,

  \quad $k$: the amount of partitions

  \textbf{Output:}

  \quad meta-data: a bag of examples
    $\Lbag z^{\prime}_1,\dots,z^{\prime}_n \Rbag$

  \begin{algorithmic}[1]
    \STATE{split bag into $k$ roughly equal sized
           partitions $\text{bag}_1,\dots,\text{bag}_k$}
    \FORALL{$\text{bag}_i, i=1,\dots,k$}
      \STATE{combine all $\text{bags} \neq \text{bag}_i$
             to the training set}
      \STATE{train $B$ with the training set}
      \STATE{let $B$ predict examples in $\text{Bag}_i$}
      \FORALL{$(x_j,y_j) \in \text{bag}_i$}
        \STATE{add element to meta data:
          $
            \Bigg(x_j, y^{\prime}_j :=
            \begin{cases}
              0 &\quad \text{if prediction of $B$ for
                             $x_j \neq y_j$}\\
              1 &\quad \text{if prediction of $B$ for
                             $x_j = y_j$}\\
            \end{cases}
            \Bigg)
          $
        }
      \ENDFOR
    \ENDFOR
    \RETURN{meta data}
  \end{algorithmic}
\end{algorithm}

\subsection*{Notes on the implementation}

\libconform{} provides the \texttt{Meta} class for
combined classifiers. In order to offer the most
flexibility for developers, \texttt{Meta} only takes two
interfaces to each classifier $B$ and $M$, one being a
function used for training the classifier, the other being
for predicting. This way \texttt{Meta} does not take the
classifiers itself, so it can be used with any other
library implementing $B$. The interfaces to $M$ could
change in future versions, so \texttt{Meta} uses $M$
directly. Currently this is not the case since using
conformal prediction and inductive conformal prediction
takes different approaches.

The ROCCH is constructed using the scipy library's
implementation (based on the qhull library) of the
Quickhull algorithm \citep[see][]{scipy,barber_et_al_1996}.

\section{Conclusion}

Like stated in the introduction, \libconform{} \version{}
is not yet stable. The API is likely to change and improve
and the library needs more tests. Furthermore there are
some algorithms from the conformal prediction framework
still missing, including:

\begin{itemize}

  \item cross-conformal prediction
        \citep[see][]{vovk_2012}

  \item aggregated conformal prediction
        \citep[see][]{carlsson_et_al_2014}

  \item inductive ridge regression confidence machine
        \citep[see][]{papadopoulos_et_al_2002}

  \item Venn abers prediction
        \citep[see][]{vovk_et_al_2014}

  \item more nonconformity scores which come out of the box
        with \libconform{}

\end{itemize}
Also the performance and especially multi-threading is an
issue which will be dealt with in future versions.

All that said, currently \libconform{} provides a very
extensible API and already implements some of the main
algorithms of the conformal prediction framework.

The goal for \libconform{} is to be one day
the go-to implementation of the conformal prediction
framework for Python. Hopefully it can attract newcomers to
use conformal prediction for their needs and will become a
community project, which grows and improves constantly.

\renewcommand{\appendixpagename}{}
\begin{appendices}
  \section*{Appendices}

  \section{API reference}
  \label{appendix:a}

    % ncs {{{
    \subsection*{\texttt{ncs}}

      Module for nonconformity measures, both for
      classification and regression.
      Instances of nonconformity measures are needed by
      \texttt{CP, ICP} and \texttt{RRCM} in order for them
      to be able to predict.
      \\

      Members

      \begin{itemize}

        % base {{{
        \item
          \texttt{ncs.base}

          Module containing the base classes from which
          nonconformity measurement implementations
          inherit.
          \\

          Members

          \begin{itemize}

            % NCSBase {{{
            \item
              \texttt{ncs.base.NCSBase}
              \label{itm:ncsbase}

              Base class for nonconformity measures for
              classification.
              If a Python object is passed to \texttt{CP}
              or \texttt{ICP} as a nonconformity measure
              and the object does not inherit from
              \texttt{NCSBase} an exception is raised.
              \\

              Methods

              \begin{itemize}

                % train {{{
                \item
                  \texttt{train(X, y)}
                  \label{itm:ncsbase_train}

                  Dummy method which needs to be
                  implemented by the inheriting
                  nonconformity measure.
                  \\

                  This method is for training the
                  underlying algorithm $D$ based on which
                  the nonconformity measure generates the
                  nonconformity scores.
                  \\

                  \begin{tabu}{llX}
                    Parameters: &\texttt{X}
                                &matrix containing the
                                 observations of a training
                                 set.
                                 \\
                                &\texttt{y}
                                &vector containing the
                                 labels of a training set.
                                 \\
                  \end{tabu}
                  \wo
                % }}}

                % scores {{{
                \item
                  \texttt{scores(X, y, cp)}

                  Dummy method which needs to be
                  implemented by the inheriting
                  nonconformity measure.
                  \\

                  Computes the nonconformity scores for
                  each example from a bag.
                  \\

                  \begin{tabu}{llX}
                    Parameters: &\texttt{X}
                                &matrix containing the
                                 observations of a bag of
                                 examples.
                                 \\
                                &\texttt{y}
                                &vector containing the
                                 labels of a bag of
                                 examples.
                                 \\
                                &\texttt{cp}
                                &boolean whether
                                 \texttt{CP} called this
                                 method or not.
                                 If \texttt{CP} called this
                                 method than \texttt{X} and
                                 \texttt{y} are equal to
                                 the training set provided
                                 to \texttt{train}.
                                 This way a nonconformity
                                 measure can implement
                                 (\ref{eq:a0}) rather than
                                 (\ref{eq:a1}).
                                 \\\\
                    Returns:    &\texttt{S}
                                &a vector with the score
                                 for each example in the
                                 bag.
                                 \\
                  \end{tabu}
                  \wo
                % }}}

                % score {{{
                \item
                  \texttt{score(x, labels)}

                  Dummy method which needs to be
                  implemented by the inheriting
                  nonconformity measure.
                  Could change in a future version from
                  the observation \texttt{x} to a bag of
                  observations \texttt{X}.
                  \\

                  This method computes the nonconformity
                  score for each new example
                  $z_g := (x_g, y), y \in \textbf{Y}$.
                  \\

                  \begin{tabu}{llX}
                    Parameters: &\texttt{x}
                                &an observation.
                                 \\
                                &\texttt{labels}
                                &all possible elements of
                                 $\textbf{Y}$.
                                 \\\\
                    Returns:    &\texttt{SL}
                                &vector with the score
                                 for each $z_g$.
                                 \\
                    \end{tabu}
                    \wo
                % }}}
              \end{itemize}
            % }}}

            % NCSBaseRegressor {{{
            \item
              \texttt{ncs.base.NCSBaseRegressor}
              \label{itm:ncsbaseregressor}

              Base class for nonconformity measures for
              regression. If a Python object is passed to
              \texttt{RRCM} as a nonconformity measure and
              the object does not inherit from
              \texttt{NCSBaseRegressor} an exception is
              raised.
              \\

              Methods

              \begin{itemize}

                \item
                  \texttt{train(X, y)}

                  Dummy method which needs to be
                  implemented by the inheriting
                  nonconformity measure.
                  \\

                  This method is for training the
                  underlying algorithm $D$ based on which
                  the nonconformity measure generates the
                  nonconformity scores.
                  \\

                  \begin{tabu}{llX}
                    Parameters: &\texttt{X}
                                &matrix containing the
                                 observations of a training
                                 set.
                                 \\
                                &\texttt{y}
                                &vector containing the
                                 labels of a training set.
                                 \\
                  \end{tabu}
                  \wo

                \item
                  \texttt{coeffs(X, y, cp)}

                  Dummy method which needs to be
                  implemented by the inheriting
                  nonconformity measure.
                  \\

                  Computes $a$ and $b$ for each example
                  from a bag.
                  \\

                  \begin{tabu}{llX}
                    Parameters: &\texttt{X}
                                &matrix containing the
                                 observations of a bag of
                                 examples.
                                 \\
                                &\texttt{y}
                                &vector containing the
                                 labels of a bag of
                                 examples.
                                 \\
                                &\texttt{cp}
                                &boolean whether
                                 \texttt{CP} called this
                                 method or not.
                                 If \texttt{CP} called this
                                 method than \texttt{X} and
                                 \texttt{y} are equal to
                                 the training set provided
                                 to \texttt{train}.
                                 This way a nonconformity
                                 measure can implement
                                 (\ref{eq:a0}) rather than
                                 (\ref{eq:a1}).
                                 \\\\
                      Returns:  &\texttt{A, B}
                                &two vectors with the $a$
                                 and $b$ for each example
                                 in the bag.
                                 \\
                    \end{tabu}
                    \wo

                \item
                  \texttt{coeffs\_n(x)}

                  Dummy method which needs to be
                  implemented by the inheriting
                  nonconformity measure.
                  Could change in a future version from the
                  observation \texttt{x} to a bag of
                  observations \texttt{X}.
                  \\

                  Computes $a$ and $b$ for a
                  new observation which should be
                  predicted.
                  \\

                  \begin{tabu}{llX}
                    Parameters: &\texttt{x}
                                &an observation.
                                 \\\\
                    Returns:    &\texttt{a, b}
                                &the coefficients $a$ and
                                 $b$ of the observation.
                                 \\
                  \end{tabu}
                  \wo

              \end{itemize}
            % }}}

          \end{itemize}
        % }}}

        % NCSDecisionTree {{{
        \item
          \texttt{ncs.NCSDecisionTree(**sklearn)}

          Class implementing a nonconformity measure based
          on a decision tree for classification.
          The score for an example $z := (x,y)$ is the
          amout of examples with the same label $y$ in the
          tree node containing $z$ devided through all the
          examples the tree node contains
          \citep[see][Chapter 4]{alrw}.
          \\

          The implementation is based on the scikit-learn
          implementation \citep[see][]{sklearn_api}.
          \\

          \begin{tabu}{llX}
            Paramters: &\texttt{**sklearn}
                       &keyword arguments for the decision
                        tree implementation of
                        scikit-learn.
          \end{tabu}
          \wo

          Methods

          See~\hyperref[itm:ncsbase]
          {\texttt{ncs.base.NCSBase}}.
          \\

        % }}}

        % NCSKNearestNeighbors {{{
        \item
          \texttt{ncs.NCSKNearestNeighbors(**sklearn)}

          Class implementing a nonconformity measure based
          on the $k$-nearest neighbors method for
          classification.
          The score for an example $z := (x,y)$ is computed
          as $\frac{d^{y}_k}{d^{-y}_k}$
          (see Chapter~\ref{subsec:ncs}).
          \\

          The implementation is based on the scikit-learn
          implementation \citep[see][]{sklearn_api}.
          \\

          \begin{tabu}{llX}
            Paramters: &\texttt{**sklearn}
                       &keyword arguments for the
                        $k$-nearest neigbors implementation
                        of scikit-learn.
          \end{tabu}
          \wo

          Methods

          See~\hyperref[itm:ncsbase]
          {\texttt{ncs.base.NCSBase}}.
          \\

        % }}}

        % NCSKNearestNeighborsRegressor {{{
        \item
          \texttt{ncs.NCSKNearestNeighborsRegressor(%
                  **sklearn)}

          Class implementing a nonconformity measure based
          on the $k$-nearest neighbors method for
          regression (see Chapter~\ref{subsec:rrcm}).
          \\

          The implementation is based on the scikit-learn
          implementation \citep[see][]{sklearn_api}.
          \\

          \begin{tabu}{llX}
            Paramters: &\texttt{**sklearn}
                       &keyword arguments for the
                        $k$-nearest neigbors implementation
                        of scikit-learn.
          \end{tabu}
          \wo

          Methods

          See~\hyperref[itm:ncsbaseregressor]
          {\texttt{ncs.base.NCSBaseRegressor}}.
          \\

        % }}}

        % NCSNeuralNet {{{
        \item
          \texttt{ncs.NCSNeuralNet(train\_, predict\_,
            scorer = "sum", gamma = 0.0)}

          Class implementing a nonconformity measure based
          on a neural net for classification.
          The nonconformity score is computed via the
          output neurons and how their outputs defer.
          For more on nonconformity scores based on neural
          nets see \citet[Chapter 4]
          {papadopoulos_et_al_2007,alrw}.
          \\

          \begin{tabu}{llX}
            Paramters: &\texttt{train\_}
                       &callable taking a bag of examples
                        splitted in \texttt{X} and
                        \texttt{y}. Should provide an
                        interface to the fitting operation
                        of a neural net.
                        \\
                       &\texttt{predict\_}
                       &callable taking a vector of
                        observations \texttt{X}. Should
                        provide an interface to the
                        predict operation of a neural net
                        and return a score for each label
                        (each output neuron).
                        \\
                       &\texttt{scorer}
                       &either callable taking a vector of
                        scores predicted by the neural
                        net and the label $y$ of the
                        example $z := (x,y)$ or a string.
                        Returns the score for $z$.
                        If \texttt{scorer} is a string it
                        has to be either \texttt{"sum",
                        "diff"} or \texttt{"max"}, each a
                        predifined nonconformity measure
                        \citep[see][Chapter 4]
                        {papadopoulos_et_al_2007,alrw}.
                        \\
                       &\texttt{gamma}
                       &constant for calibrating the
                        \texttt{"sum"} and the
                        \texttt{"max"} nonconformity
                        measures.
                        Ignored if \texttt{scorer} is not
                        \texttt{"sum"} or \texttt{"max"}.
                        \\
          \end{tabu}
          \wo

          Methods

          See~\hyperref[itm:ncsbase]
          {\texttt{ncs.base.NCSBase}}.
        % }}}

      \end{itemize}
    % }}}

    % vtx {{{
    \subsection*{\texttt{vtx}}

      Module for Venn taxonomies. Will be deprecated in
      future versions in favor of a module for
      $n$-taxonomies (see
      Chapters~\ref{sec:mcp},~\ref{sec:venn}).
      \\

      Members

      \begin{itemize}

        % base {{{
        \item
          \texttt{vtx.base}

          Module containing the base class from which Venn
          taxonomies implementations inherit.
          \\

          Members

          \begin{itemize}

            % VTXBase {{{
            \item
              \texttt{vtx.base.VTXBase}
              \label{itm:vtxbase}

              Base class for Venn taxonomies. If a Python
              object is passed to \texttt{Venn} as a
              Venn taxonmy and the object does not inherit
              from \texttt{VTXBase} an exception is raised.
              \\

              Methods

              \begin{itemize}

                % train {{{
                \item
                  \texttt{train(X, y)}

                  Dummy method which needs to be
                  implemented by the inheriting
                  Venn taxonomy.
                  \\

                  This method is for training the
                  underlying algorithm $D$ based on which
                  a Venn taxonomy generates its categories.
                  \\

                  \begin{tabu}{llX}
                    Parameters: &\texttt{X}
                                &matrix containing the
                                 observations of a training
                                 set.
                                 \\
                                &\texttt{y}
                                &vector containing the
                                 labels of a training set.
                                 \\
                  \end{tabu}
                  \wo
                % }}}

                % category {{{
                \item
                  \texttt{category(x, y, contains\_x)}

                  Dummy method which needs to be
                  implemented by the inheriting
                  Venn taxonomy.
                  \\

                  Computes the category of the example
                  $z := (\texttt{x}, \texttt{y})$.
                  \\

                  \begin{tabu}{llX}
                    Parameters: &\texttt{x}
                                &an observation.
                                 \\
                                &\texttt{y}
                                &a label.
                                 \\
                                &\texttt{contains\_x}
                                &boolean whether
                                $x$ was part of the
                                training set passed to the
                                \texttt{train} method.
                                Some Venn taxonomies may
                                want to change their
                                behaviour if $x$ was part
                                of the training set.
                                 \\\\
                    Returns:    &\texttt{c}
                                &a category for the
                                 example $z$.
                                 \\
                  \end{tabu}
                  \wo
                % }}}

              \end{itemize}
            % }}}

          \end{itemize}
        % }}}

        % VTXKNearestNeighbors {{{
        \item
          \texttt{vtx.VTXKNearestNeighbors(**sklearn)}

          Class implementing a Venn taxonomy based
          on the $k$-nearest neighbors method for
          classification (see Chapter~\ref{sec:venn}).
          \\

          The implementation is based on the scikit-learn
          implementation \citep[see][]{sklearn_api}.
          \\

          \begin{tabu}{llX}
            Paramters: &\texttt{**sklearn}
                       &keyword arguments for the
                        $k$-nearest neigbors implementation
                        of scikit-learn.
          \end{tabu}
          \wo

          Methods

          See~\hyperref[itm:vtxbase]
          {\texttt{vtx.base.VTXBase}}.
          \\

        % }}}

      \end{itemize}
    % }}}

    % CP {{{
    \subsection*{\texttt{CP(A, epsilons, smoothed = False,
        mondrian\_taxonomy = \_not\_mcp)}}
      \label{itm:cp}

      Class implementing conformal prediction
      (see Chapter~\ref{sec:cp}).
      \\

      \begin{tabu}{llX}
        Parameters: &\texttt{A}
                    &instance of a nonconformity measure
                     (Python object inheriting from
                     \hyperref[itm:ncsbase]
                     {\texttt{ncs.base.NCSBase}}).
                     \\
                    &\texttt{epsilons}
                    &list containing significance levels.
                     \\
                    &\texttt{smoothed}
                    &flag whether the predictor should use
                     (\ref{eq:p1}) rather than
                     (\ref{eq:p0}) for computing the
                     p-value of an example.
                     \\
                   &\texttt{mondrian\_taxonomy}
                   &callable for a $1$-taxonomy making the
                    predictor a MCP
                    (see Chapter~\ref{sec:mcp}).
                    The default \texttt{\_not\_mcp} does
                    not impose any taxonomy on the
                    predictor, mapping each example to the
                    same category.
                    \\
      \end{tabu}

      Methods

      \begin{itemize}

        % train {{{
        \item
          \texttt{train(X, y, override = False)}
          \label{itm:cp_train}

          Wrapper for the train method of the nonconformity
          measure \texttt{A} (see \\ \hyperref
          [itm:ncsbase_train]
          {\texttt{ncs.base.NCSBase.train}}).

          \begin{tabu}{llX}
            Parameters: &\texttt{X}
                        &matrix containing the observations
                         of a training set.
                         \\
                        &\texttt{y}
                        &vector containing the labels of a
                         training set.
                         \\
                        &\texttt{override}
                        &flag whether the training set
                         should be appended to the training
                         set already given to the
                         \texttt{CP} instance or override
                         it.
          \end{tabu}
          \wo
        % }}}

        % predict {{{
        \item
          \texttt{predict(X)}
          \label{itm:cp_predict}

          \begin{tabu}{llX}
            Paramters: &\texttt{X}
                       &matrix containing observations
                        which should be predicted.
                        \\\\
            Returns:   &\texttt{predictions}
                       &the predictions for the
                        observations provided by
                        \texttt{X}.
          \end{tabu}
          \wo
        % }}}

        % predict_best {{{
        \item
          \texttt{predict\_best(X,
                  significance\_levels = True)}
          \label{itm:cp_predict_best}

          Method for determining only the best label as
          prediction.
          \\

          \begin{tabu}{lp{4cm}X}
            Parameters: &\texttt{X}
                        &matrix containing observations
                         which should be predicted.
                         \\
                        &\texttt{significance\_levels}
                        &flag whether the significance
                         levels of the predictions should
                         be returned, too.
                         \\\\
            Returns:    &\texttt{predictions,
                                 sig\_levels}
                        &a vector with the best labels,
                         optionally a vector with the
                         significance level of the best
                         label.
          \end{tabu}
          \wo
        % }}}

        % score {{{
        \item
          \texttt{score(X, y)}
          \label{itm:cp_score}

          Generates metrics for the predictor.
          \\

          \begin{tabu}{llX}
            Parameters: &\texttt{X}
                        &matrix containing observations
                         which should be predicted.
                         \\
                        &\texttt{y}
                        &vector with the corresponding
                         labels to the observations
                         provided by \texttt{X}.
                         \\\\
            Returns:    &\texttt{metrics}
                        &metrics for the predictor.
          \end{tabu}
          \wo
        % }}}

        % score_online {{{
        \item
          \texttt{score\_online(X, y)}
          \label{itm:cp_score_online}

          Iterates through a set of examples, predicts each
          example and then trains the predictor with it
          before predicting the next example. Generates
          metrics for the predictor.
          \\

          \begin{tabu}{llX}
            Parameters: &\texttt{X}
                        &matrix containing observations
                         which should be predicted.
                         \\
                        &\texttt{y}
                        &vector with the corresponding
                         labels to the observations
                         provided by \texttt{X}.
                         \\\\
            Returns:    &\texttt{metrics}
                        &metrics for the predictor.
          \end{tabu}
          \wo
        % }}}

        % p_vals {{{
        \item
          \texttt{p\_vals(X)}
          \label{itm:cp_p_vals}

          Computes the p-values for each observation from
          a bag, paired with each label $y \in \textbf{Y}$.
          \\

          \begin{tabu}{llX}
            Parameters: &\texttt{X}
                        &matrix containing observations
                         for which the p-values are
                         generated.
                         \\\\
            Returns:    &\texttt{p\_vals}
                        &a matrix of p-values for each
                         observation from \texttt{X} paired
                         with each label
                         $y \in \textbf{Y}$.
          \end{tabu}
        % }}}

      \end{itemize}
    % }}}

    % ICP {{{
    \subsection*{\texttt{ICP(A, epsilons, smoothed = False,
        mondrian\_taxonomy = \_not\_mcp)}}

      Class implementing inductive conformal prediction
      (see Chapter~\ref{sec:icp}).
      \\

      \begin{tabu}{llX}
        Parameters: &&see~\hyperref[itm:cp]{\texttt{CP}}.
      \end{tabu}
      \wo
      Methods

      \begin{itemize}

        % train {{{
        \item
          \texttt{train(X, y, override = False)}

          See~\hyperref[itm:cp_train]{\texttt{CP.train}}.
          \\
        % }}}

        % calibrate {{{
        \item
          \texttt{calibrate(X, y, override = False)}

          Method for calibrating the ICP (see
          Chapter~\ref{sec:icp}).
          \\

          \begin{tabu}{llX}
            Parameters: &\texttt{X}
                        &matrix containing the observations
                         of a calibration set.
                         \\
                        &\texttt{y}
                        &vector containing the labels of
                         a calibration set.
                         \\
                        &\texttt{override}
                        &flag whether the calibration set
                         should be appended to the
                         calibration set already given to
                         the \texttt{ICP} instance or
                         should override it.
          \end{tabu}
          \wo
        % }}}

        % predict {{{
        \item
          \texttt{predict(X)}

          See~\hyperref[itm:cp_predict]
          {\texttt{CP.predict}}.
          \\
        % }}}

        % predict_best {{{
        \item
          \texttt{predict\_best(X, p\_vals = True)}

          See~\hyperref[itm:cp_predict_best]
          {\texttt{CP.predict\_best}}.
          \\
        % }}}

        % score {{{
        \item
          \texttt{score(X, y)}

          See~\hyperref[itm:cp_score]
          {\texttt{CP.score}}.
          \\
        % }}}

        % p_vals {{{
        \item
          \texttt{p\_vals(X)}

          See~\hyperref[itm:cp_p_vals]
          {\texttt{CP.p\_vals}}.
        % }}}

      \end{itemize}
    % }}}

    % RRCM {{{
    \subsection*{\texttt{RRCM(A, epsilons,
                 convex\_hull = True)}}

      Class implementing the ridge regression confidence
      machine (see Chapter~\ref{subsec:rrcm}).
      \\

      \begin{tabu}{llX}
        Parameters: &\texttt{A}
                    &instance of a nonconformity
                     measure (Python object inheriting from
                     \hyperref[itm:ncsbaseregressor]
                     {\texttt{ncs.base.NCSBaseRegressor}}).
                     \\
                    &\texttt{epsilons}
                    &list containing significance levels.
                     \\
                    &\texttt{convex\_hull}
                    &flag whether the \texttt{RRCM}
                     instance should return the convex hull
                     as the prediction interval (see
                     Chapter~\ref{subsec:rrcm}).
      \end{tabu}

      Methods

      \begin{itemize}

        % train {{{
        \item
          \texttt{train(X, y, override = False)}

          See~\hyperref[itm:cp_train]
          {\texttt{CP.train}}.
          \\
        % }}}

        % predict {{{
        \item
          \texttt{predict(X)}

          See~\hyperref[itm:cp_predict]
          {\texttt{CP.predict}}.
          \\
        % }}}

        % score {{{
        \item
          \texttt{score(X, y)}

          See~\hyperref[itm:cp_score]
          {\texttt{CP.score}}.
          \\
        % }}}

        % score_online {{{
        \item
          \texttt{score\_online(X, y)}

          See~\hyperref[itm:cp_score_online]
          {\texttt{CP.score\_online}}.
        % }}}

      \end{itemize}
    % }}}

    % Venn {{{
    \subsection*{\texttt{Venn(venn\_taxonomy)}}

      Class implementing Venn prediction
      (see Chapter~\ref{sec:venn}).
      \\

      \begin{tabu}{llX}
        Parameters: &\texttt{venn\_taxonomy}
                    &instance of a Venn taxonomy (Python
                     object inheriting from
                     \hyperref[itm:vtxbase]
                     {\texttt{vtx.base.VTXBase}}).
      \end{tabu}

      Methods

      \begin{itemize}

        % train {{{
        \item
          \texttt{train(X, y, override = False)}

          See~\hyperref[itm:cp_train]
          {\texttt{CP.train}}.
          \\
        % }}}

        % predict {{{
        \item
          \texttt{predict(X, proba = True)}
          \label{itm:cp_predict}

          \begin{tabu}{lp{4cm}X}
            Paramters: &\texttt{X}
                       &matrix containing observations
                        which should be predicted.
                        \\
                       &\texttt{proba}
                       &flag whether the error probability
                        interval of the predictions should
                        be returned, too.
                        \\\\
            Returns:   &\texttt{predictions,
                                error\_probabilites}
                       &a vector with the predicted labels,
                        optionally a vector with the
                        error probability intervals.
          \end{tabu}
          \wo
        % }}}

        % score {{{
        \item
          \texttt{score(X, y)}

          See~\hyperref[itm:cp_score]
          {\texttt{CP.score}}.
          \\
        % }}}

        % score_online {{{
        \item
          \texttt{score\_online(X, y)}

          See~\hyperref[itm:cp_score_online]
          {\texttt{CP.score\_online}}.
        % }}}

      \end{itemize}
    % }}}

    % Meta {{{
    \subsection*{\texttt{Meta(M\_train, M\_predict,
                 B\_train, B\_predict, epsilons)}}

      Class implementing meta-conformal prediction
      (see Chapter~\ref{sec:meta}).
      \\

      \begin{tabu}{llX}
        Parameters: &\texttt{M\_train}
                    &callable interfacing with the meta
                     classifier for training it.
                     \\
                    &\texttt{M\_predict}
                    &callable interfacing with the meta
                     classifier for predicting.
                     \\
                    &\texttt{B\_train}
                    &callable interfacing with the base
                     classifier for training it.
                     \\
                    &\texttt{B\_predict}
                    &callable interfacing with the bas
                     classifier for predicting.
                     \\
                    &\texttt{epsilons}
                    &list containing significance levels.
      \end{tabu}

      Methods

      \begin{itemize}

        % train {{{
        \item
          \texttt{train(X, y, k\_folds, plot = False)}

          Generates the internal threshold $T$ and trains
          $B$ and $M$.
          \\

          \begin{tabu}{llX}
            Parameters: &\texttt{X}
                        &matrix containing the observations
                         of a training set.
                         \\
                        &\texttt{y}
                        &vector containing the labels of a
                         training set.
                         \\
                        &\texttt{k\_folds}
                        &how many partitions are use for
                         generating the meta data and
                         scores.
                         \\
                        &\texttt{plot}
                        &flag whether a matplotlib
                         \citep[see][]{hunter_2007} is
                         shown with the ROC curve, the
                         ROCCH and the iso-precision lines.
                         If the plot is shown the execution
                         of \texttt{train} is stopped.
          \end{tabu}
        % }}}

        % predict {{{
        \item
          \texttt{predict(X)}

          See~\hyperref[itm:cp_predict]
          {\texttt{CP.predict}}.
          \\
        % }}}

        % score {{{
        \item
          \texttt{score(X, y)}

          See~\hyperref[itm:cp_score]
          {\texttt{CP.score}}.
          \\
        % }}}

      \end{itemize}
    % }}}

  \lstset{%
    basicstyle=\footnotesize,
    numbers = left,
    firstnumber = 0,
    stepnumber = 5,
    frame = single,
    breaklines = true,
  }

  \section{Examples}
  \label{appendix:b}

  \subsection{\texttt{CP} on the iris data set with
              different nonconformity measures}

\begin{lstlisting}[language=Python]
import numpy as np

from libconform import CP
from libconform.ncs import NCSDecisionTree, NCSKNearestNeighbors

from sklearn.datasets import load_iris

X, y = load_iris(True)

# randomly permute X, y
indices = np.arange(len(X))
np.random.shuffle(indices)

X = X[indices]
y = y[indices]

X_train, y_train = X[:-20], y[:-20]
X_test,  y_test  = X[-20:], y[-20:]

epsilons = [0.01,0.025,0.05,0.1]

# offline decision tree
ncs = NCSDecisionTree()
cp = CP(ncs, epsilons)

cp.train(X_train, y_train)
res = cp.score(X_test, y_test)
print(res)

# online decision tree
ncs = NCSDecisionTree()
cp = CP(ncs, epsilons)

res = cp.score_online(X, y)
print(res)

# offline nearest neighbors
ncs = NCSKNearestNeighbors(n_neighbors=1)
cp = CP(ncs, epsilons)

cp.train(X_train, y_train)
res = cp.score(X_test, y_test)
print(res)

# online nearest neighbors
ncs = NCSKNearestNeighbors(n_neighbors=1)
cp = CP(ncs, epsilons)

cp.score_online(X, y)
print(res)
\end{lstlisting}

  \subsubsection{Nonconformity measure based on keras
                 neural net}

\begin{lstlisting}[language=Python]
import numpy as np

from keras.layers import Dense
from keras.models import Sequential

from libconform import CP
from libconform.ncs import NCSNeuralNet

from sklearn.datasets import load_iris

def compile_model(in_dim, out_dim):

    # keras
    model = Sequential()

    model.add(Dense( units=10, activation='tanh'
                   , input_dim=in_dim ))
    model.add(Dense(units=10, activation='tanh'))
    model.add(Dense( units=out_dim
                   , activation='softmax'))

    model.compile(
        loss='categorical_crossentropy',
        optimizer='adam',
        metrics=['accuracy']
    )

    return model

epsilons = [0.01,0.025,0.05,0.1]

X, y = load_iris(True)

labels = np.unique(y)

y = np.array([[0. if j != v else 1.0 for j in labels] for v in y])

# randomly permute X, y
indices = np.arange(len(X))
np.random.shuffle(indices)

X = X[indices]
y = y[indices]

X_train, y_train = X[:-25], y[:-25]
X_test,  y_test  = X[-25:], y[-25:]

model = compile_model(X.shape[1], y.shape[1])

train = lambda X, y: model.fit(X, y, epochs=5)
predict = lambda X: model.predict(X)

# first training on the majority of data, then the last 25 examples online
ncs = NCSNeuralNet(train, predict)
cp = CP(ncs, epsilons)

cp.train(X_train, y_train)
res = cp.score_online(X_test, y_test)
print(res)
\end{lstlisting}

  \subsection{\texttt{ICP} on the iris data set with
              different nonconformity measures}

\begin{lstlisting}[language=Python]
import numpy as np

from libconform import ICP
from libconform.ncs import NCSDecisionTree, NCSKNearestNeighbors

from sklearn.datasets import load_iris

X, y = load_iris(True)

# randomly permute X, y
indices = np.arange(len(X))
np.random.shuffle(indices)

X = X[indices]
y = y[indices]

X_train, y_train = X[:-50],    y[:-50]
X_cal,   y_cal   = X[-50:-20], y[-50:-20]
X_test,  y_test  = X[-20:],    y[-20:]

epsilons = [0.01,0.025,0.05,0.1]

# decision tree
ncs = NCSDecisionTree()
icp = ICP(ncs, epsilons)

icp.train(X_train, y_train)
icp.calibrate(X_cal, y_cal)
res = icp.score(X_test, y_test)
print(res)

# nearest neighbors
ncs = NCSKNearestNeighbors(n_neighbors=1)
icp = ICP(ncs, epsilons)

icp.train(X_train, y_train)
icp.calibrate(X_cal, y_cal)
res = icp.score(X_test, y_test)
print(res)
\end{lstlisting}

  \subsection{\texttt{MCP} on the iris data set, both
              inductive and transductive}

\begin{lstlisting}[language=Python]
import numpy as np

from libconform import CP, ICP
from libconform.ncs import NCSKNearestNeighbors

from sklearn.datasets import load_iris

X, y = load_iris(True)

# randomly permute X, y
indices = np.arange(len(X))
np.random.shuffle(indices)

X = X[indices]
y = y[indices]

epsilons = [0.01,0.025,0.05,0.1]

def label_conditional(observation, label):
    return label

# mondrian conformal prediction
X_train, y_train = X[:-20],    y[:-20]
X_test,  y_test  = X[-20:],    y[-20:]

ncs = NCSKNearestNeighbors(n_neighbors=1)
cp = CP(ncs, epsilons, mondrian_taxonomy=label_conditional)

cp.train(X_train, y_train)
res = cp.score_online(X_test, y_test)
print(res)

# mondrian inductive conformal prediction
X_train, y_train = X[:-50],    y[:-50]
X_cal,   y_cal   = X[-50:-20], y[-50:-20]

ncs = NCSKNearestNeighbors(n_neighbors=1)
icp = ICP(ncs, epsilons, mondrian_taxonomy=label_conditional)

icp.train(X_train, y_train)
icp.calibrate(X_cal, y_cal)
res = icp.score(X_test, y_test)
print(res)
\end{lstlisting}

  \subsection{\texttt{RRCM} on the Boston housing data set,
              both online and offline}

\begin{lstlisting}[language=Python]
import numpy as np

from libconform import RRCM
from libconform.ncs import NCSKNearestNeighborsRegressor

from sklearn.datasets import load_boston

X, y = load_boston(True)

# randomly permute X, y
indices = np.arange(len(X))
np.random.shuffle(indices)

X = X[indices]
y = y[indices]

X_train, y_train = X[:-20],    y[:-20]
X_test,  y_test  = X[-20:],    y[-20:]

epsilons = [0.01,0.025,0.05,0.1]

# online
ncs  = NCSKNearestNeighborsRegressor(n_neighbors=1)
rrcm = RRCM(ncs, epsilons)

res = rrcm.score_online(X, y)
print(res)

# offline
ncs  = NCSKNearestNeighborsRegressor(n_neighbors=1)
rrcm = RRCM(ncs, epsilons)

rrcm.train(X_train, y_train)
res = rrcm.score(X_test, y_test)
print(res)
\end{lstlisting}

  \subsection{\texttt{Venn} on the iris data set}

\begin{lstlisting}[language=Python]
import numpy as np

from libconform import Venn
from libconform.vtx import VTXKNearestNeighbors

from sklearn.datasets import load_iris

X, y = load_iris(True)

# randomly permute X, y
indices = np.arange(len(X))
np.random.shuffle(indices)

X = X[indices]
y = y[indices]

X_train, y_train = X[:-20],    y[:-20]
X_test,  y_test  = X[-20:],    y[-20:]

vtx  = VTXKNearestNeighbors(n_neighbors=1)
venn = Venn(vtx)

venn.train(X_train, y_train)
res = venn.score(X_test, y_test)
print(res)
\end{lstlisting}

  \subsection{\texttt{Meta} on the iris data set}

\begin{lstlisting}[language=Python]
import numpy as np

from libconform import Meta, CP
from libconform.ncs import NCSKNearestNeighbors

from sklearn.datasets import load_iris
from sklearn.neighbors import KNeighborsClassifier

X, y = load_iris(True)

# randomly permute X, y
indices = np.arange(len(X))
np.random.shuffle(indices)

X = X[indices]
y = y[indices]

X_train, y_train = X[:-20],    y[:-20]
X_test,  y_test  = X[-20:],    y[-20:]

epsilons = [0.01,0.025,0.05,0.1]

B         = KNeighborsClassifier(n_neighbors=1)
B_train   = lambda X, y: B.fit(X,y)
B_predict = lambda X: B.predict(X)

ncs       = NCSKNearestNeighbors(n_neighbors=1)
M         = CP(ncs, [])
M_train   = lambda X, y: M.train(X, y)
M_predict = lambda X: M.p_vals(X)

clf = Meta(M_train, M_predict, B_train, B_predict, epsilons)

clf.train(X_train, y_train, k_folds = 10)

res = clf.score(X_test, y_test)
print(res)
\end{lstlisting}

\end{appendices}

\bibliography{libconform.bib}

\end{document}